\documentclass[letterpaper, 10 pt, journal, twoside]{IEEEtran}

\usepackage{cite}
\usepackage{booktabs}
\usepackage[pdftex]{graphicx}
\usepackage[export]{adjustbox}
\usepackage{svg}
\usepackage{tikz}
\usepackage{xcolor}
\usepackage{subcaption}
\usepackage{multirow}
\usepackage{titlesec}
\IEEEaftertitletext{\vspace{-20pt}}
\usepackage{gensymb}
\usepackage{amsmath}
\usepackage{amssymb}
\usepackage{hyperref}
\newcommand{\StatePT}{\mathcal{S}_{\text{PT}}}
\newcommand{\StateHA}{\mathcal{S}_{\text{HA}}}

\begin{document}
\title{Ratatouille: Imitation Learning Ingredients for Real-world Social Robot Navigation}

\author{James~R.~Han$^{1}$, Mithun~Vanniasinghe$^{1}$, Hshmat~Sahak$^{1}$, Nicholas Rhinehart$^{1}$, Timothy~D.~Barfoot$^{1}$

\thanks{\textsuperscript{1} University of Toronto Institute for Aerospace Studies, Canada {\tt\footnotesize \{jamesr.han, mithun.vanniasinghe, hshmat.sahak, nick.rhinehart, tim.barfoot\}@mail.utoronto.ca}}

}

\maketitle

\begin{abstract}
Scaling Reinforcement Learning to \emph{in-the-wild} social robot navigation is both data-intensive and unsafe, since policies must learn through direct interaction and inevitably encounter collisions. Offline Imitation learning (IL) avoids these risks by collecting expert demonstrations safely, training entirely offline, and deploying policies zero-shot. However, we find that na\"ively applying Behaviour Cloning (BC) to social navigation is insufficient; achieving strong performance requires careful architectural and training choices. We present \textbf{Ratatouille}, a pipeline and model architecture that, without changing the data, reduces collisions per meter by $6\times$ and improves success rate by $3\times$ compared to na\"ive BC. We validate our approach in both simulation and the real world, where we collected over 11 hours of data on a dense university campus. We further demonstrate qualitative results in a public food court. Our findings highlight that thoughtful IL design, rather than additional data, can substantially improve safety and reliability in real-world social navigation. Video: \url{https://youtu.be/tOdLTXsaYLQ}. Code will be released after acceptance.

\end{abstract}


\IEEEpeerreviewmaketitle

\section{Introduction}
\IEEEPARstart{S}{ocial} robot navigation has diverse applications, from robots transporting goods to assisting with household tasks. These settings vary widely in complexity, ranging from a single human in a household to dense pedestrian flows in shopping malls. Achieving reliable navigation in the wild remains an open challenge. While Deep Reinforcement Learning (DRL) has demonstrated promise in simulation \cite{IntentionAwareGraph, NaviSTAR, DRMPC}, its application to the real world faces two substantial obstacles. First, online DRL often requires unsafe trial-and-error interactions, leading to collisions. Second, it is difficult to specify a reward function that encodes complex social norms. Imitation Learning (IL) offers a potentially safer alternative. Expert demonstrations can be collected in complex scenarios without collisions, and the learned policy implicitly captures the expert’s social norms and internal reward function. IL has been successfully applied across domains, from early lane-following systems \cite{ALVINN} to recent advances in mobile manipulation with vision-language-action models \cite{pi05}. However, the effectiveness of Behaviour Cloning (BC) remains inconsistent: while it performs well in some settings, it fails dramatically in others \cite{causalconfusion, copycatBC}. In our studies, na\"ive BC produces poor policies for social navigation, requiring careful design choices for BC to produce proficient policies. 

In this paper, we introduce \textit{Ratatouille}, a BC recipe that substantially enhances navigation performance. Ratatouille integrates several key ingredients—\emph{weight perturbations}, \emph{data augmentation}, \emph{a hybrid action space}, \emph{action chunking}, and \emph{neural network optimization techniques}—into a unified pipeline. We conduct extensive evaluations in simulation and the real world, and ablate each component to justify its contribution. Our results show that Ratatouille achieves over a $3\times$ improvement in success rate ($28\% \rightarrow 98\%$) and a $6\times$ reduction in collisions per meter ($0.06 \rightarrow 0.01$) over na\"ive BC. We also demonstrate qualitative results in a food court, where Ratatouille exhibits proficient in-the-wild performance.

\begin{figure}
  \centering
  \includegraphics[width=0.5\textwidth]{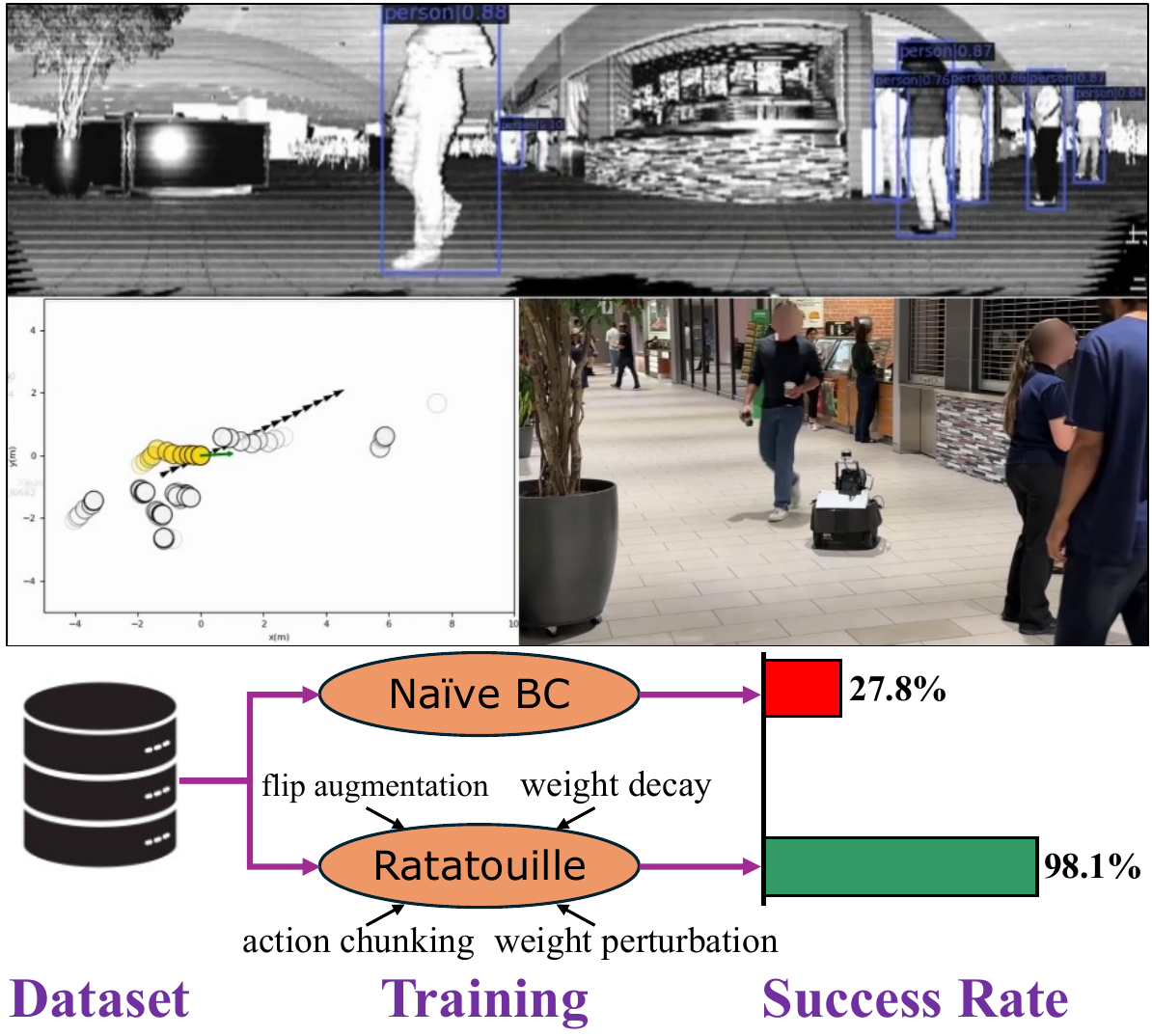} 
  \caption{Real-world evaluation in a food court. The robot navigates around a group of stationary humans and smoothly passes an oncoming pedestrian. \emph{Top:} YOLOX on LiDAR reflectivity image. \emph{Left:} Model state. \emph{Right:} Camera perspective.}
  \label{fig:RealWorldMoneyShot}
\end{figure}

\section{Background and Related Works}

\subsection{Behavioural Cloning}\label{sec:BC}
Given a dataset $D_\text{expert}$ of state-action pairs $(s, a)$ generated by an expert human driver, the goal is to train a policy $\pi$ to mimic the expert. BC matches the expert's actions through standard supervised learning. In the case of discrete actions, BC minimizes the negative log-likelihood of the expert’s actions under the policy's distribution:

\begin{equation}\label{eq:BC discrete}
L_{\text{discrete}} = \mathbb{E}_{(s,a) \sim D_\text{expert}}-\log \pi(a|s)
\end{equation}

In BC with continuous action spaces, we minimize the mean squared error (MSE) between the predicted and expert actions:
\begin{equation}\label{eq:BC continuous}
L_{\text{continuous}} = \mathbb{E}_{(s,a) \sim D_\text{expert}}||\pi(a|s) - a ||_2^2
\end{equation}

In this framework, the task objectives are encoded implicitly within the demonstrated behaviour. The desired outcome is that if the learned policy can reliably reproduce expert-like actions, the resulting robot behaviour will resemble that of the expert.

\begin{figure}[t]
    \centering
    \includegraphics[width=0.5\textwidth]{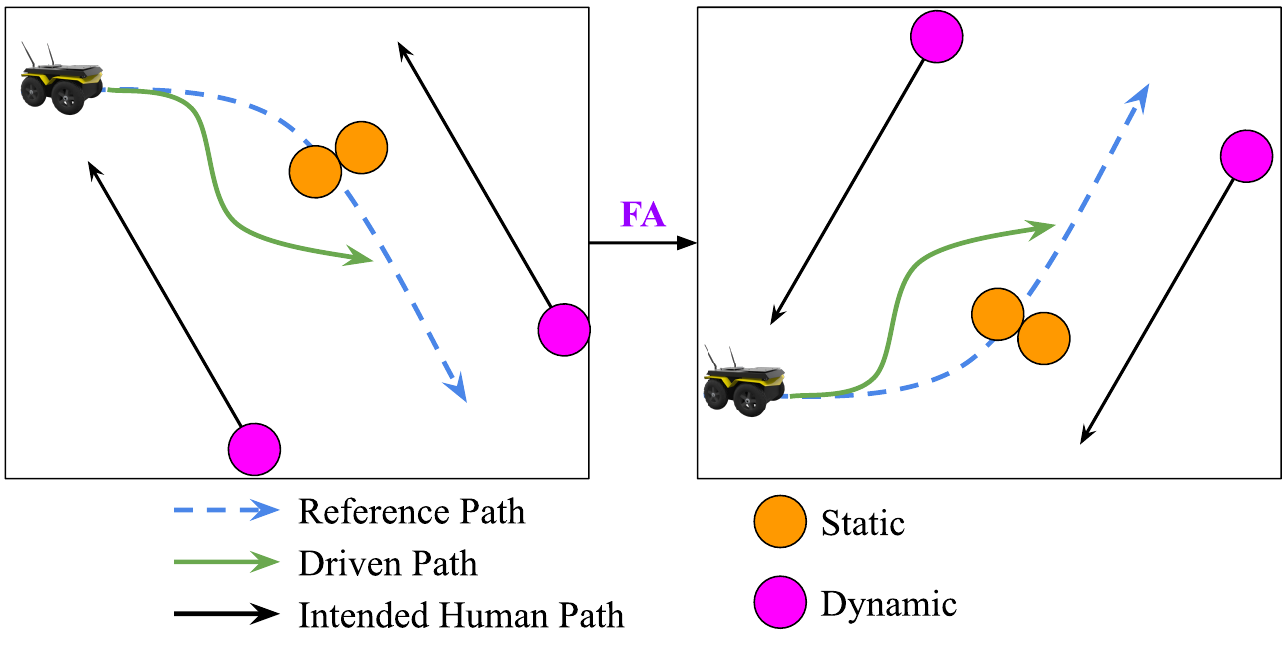}
    \caption{Flip augmentation (``FA'') example. Both scenarios are in the robot's frame. The humans, path, and expert's driven path are flipped about the robot's $x$-axis.}
    \label{fig:Flip Augmentation}
\end{figure}

\subsection{State and Action}
Our state and action formulation follows \cite{DRMPC}, which integrates path tracking with human avoidance. The action for our unicycle robot is a linear and angular velocity. Our state is divided into $\StatePT$ and $\StateHA$. A visual example is shown in Figure~\ref{fig:RealWorldMoneyShot}.  

\begin{figure*}[htbp]
    \centering
    \includegraphics[width=0.95\textwidth]{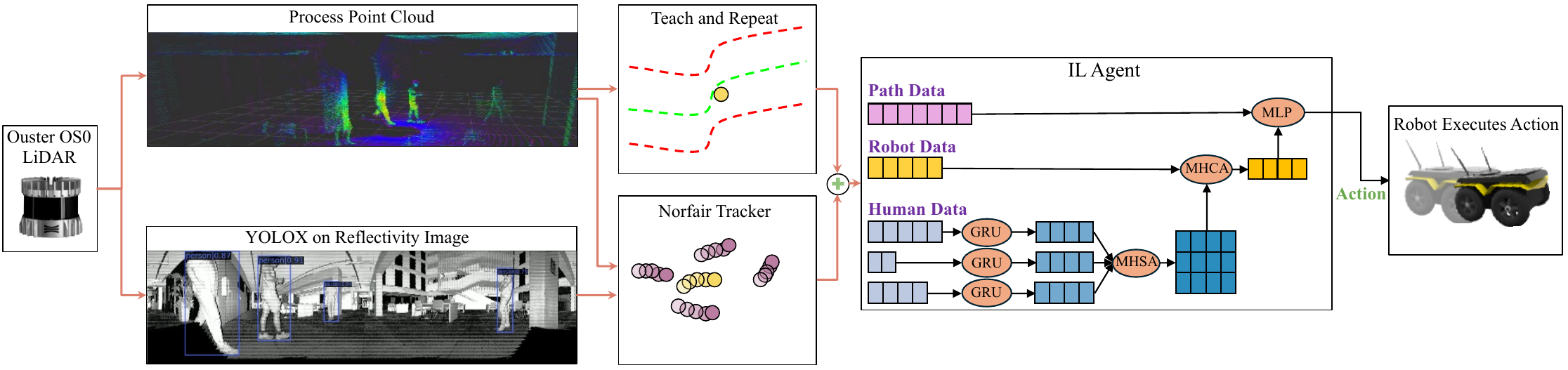}
    \caption{Real-world system from \cite{DRMPC}. Using a single LiDAR, we are able to perform both localization with T\&R, human detection with YOLOX, and human tracking with the Norfair tracker. The core architecture of the IL agent comes from \cite{DRMPC}. GRU: Gated Recurrent Unit; MHSA: Multi-Head Self Attention; MHCA: Multi-Head Cross Attention; MLP: Multilayer Perceptron. }
    \label{fig:Full Pipeline}
\end{figure*}

$\StatePT$ encodes a local segment of the path through a sequence of node poses in the robot frame. $\StateHA$ consists of the past positions of both the robot and humans in the robot frame. The number of humans and the length of their history may vary across states, but the history is capped at 10 steps. Each past position is sampled at 0.25\,s intervals, resulting in a maximum temporal horizon of 2.5\,s, which aligns well with trajectory forecasting work \cite{trajforecasting2s}. The core of our model architecture comes from \cite{DRMPC} (Figure~\ref{fig:Full Pipeline}), which elegantly handles both the varying number of humans and their variable history lengths.

\subsection{Imitation Learning in Social Robot Navigation}
Several works have adopted BC for social robot navigation \cite{SociallyCompliantNavigationThroughRawDepth, DeepMotion, DeepILWheelchair, SCAND}, differing in their choice of datasets, robot platforms, network architectures, and action representations. For instance, \cite{SociallyCompliantNavigationThroughRawDepth} performs BC directly from raw depth images, while \cite{SCAND} processes 2D LiDAR scans into bird’s-eye view (BEV) images. Due to differences in sensing and pipeline design, most approaches do not benchmark against each other, as each paper targets a distinct application context.

Most datasets used in these works were collected in real-world settings and typically span between one and ten hours of demonstration data. However, the resulting policies are often evaluated in constrained scenarios with limited human interaction \cite{SCAND, SACSoN}. Consequently, while IL can in principle encode socially compliant behaviours from demonstrations, its practical ability in diverse, real-world settings remains underexplored in the literature.  

Another branch of IL for social navigation leverages large-scale datasets originally developed for pedestrian trajectory forecasting \cite{SocialGAIL, DeepMotion}. For example, \cite{SocialGAIL} trains a policy on 12,000 human trajectories. While their results qualitatively demonstrate socially compliant trajectory prediction, the policy is not evaluated in a closed-loop control setting where the robot physically navigates among humans.  

More general-purpose datasets for social robot navigation are beginning to emerge. For example, \cite{SocialNavDatasetFromHuman} collected over 20 hours of human navigation data using a head-mounted sensor suite. The demonstrator walked in a socially compliant manner, aiming to enable cross-embodiment learning where models trained on human-centered data could generalize to robotic navigation tasks. However, this cross-embodiment learning was not attempted in their study.  

\section{Ingredients}\label{sec:related_works_ingredients}
In this section, we introduce the key ingredients for Ratatouille and their corresponding literature. 

\subsection{Data Augmentation}
Model performance in Machine Learning (ML) often scales with dataset size \cite{DataAugmentationGeneral}, thus BC performance benefits from larger datasets \cite{DataAugmentationIL}. However, data augmentation in IL presents a challenge: modifying the state often requires recomputing the corresponding expert action, unlike in classification tasks where the label remains fixed. Recomputing expert actions may be infeasible due to the effort involved, or it may be difficult to obtain a high-quality label, since driving is inherently fluid and not easily annotated at a single timestep.

For our data augmentation, we flip the state about the robot's $x$-axis (Figure \ref{fig:Flip Augmentation}) by negating the $y$-coordinates of all components (robot, humans, path) and negating the sign of the angular velocity. This transformation doubles the dataset size without degrading the quality of the expert labels. 

This flip augmentation increases both path and human configuration diversity. For example, if the original dataset contains only counter-clockwise (CCW) paths, the augmentation introduces clockwise (CW) paths. Or, if the original data shows the expert passing to the left of a static human, the augmented data introduces a mirrored scenario where the robot passes on the right. We find this easy data augmentation procedure has a large boon on performance.

\subsection{Action Chunking}
Action chunking refers to predicting a sequence of future actions instead of just the next immediate action. Several works have shown the benefits of this strategy in IL \cite{DiffusionPolicy, finetuningvisionlanguageaction}. Although the exact reason for why action chunking helps is still an open question, \cite{DiffusionPolicy} found an `inverted-U' relationship between action horizon and policy performance. We refer to the action horizon as $H$, and we find a small improvement when employing action chunking with $H=3$.

\subsection{Hybrid Action Space for Efficient Action Chunking}

BC tends to yield lower compounding errors when applied to discrete action spaces compared to continuous ones \cite{pitfallIL}. However, when the actions are originally continuous, the action discretization introduces a trade-off. A sparse discretization limits the granularity of control, but a dense discretization increases the output dimensionality, making the learning problem more complex and data-hungry. Another drawback of discrete action spaces is the statistical inefficiency of action chunking. With $n$ discrete bins, the number of action sequences grows exponentially as $n^H$. With limited data, many sequences will likely receive few or no samples.

Thus, we propose a hybrid action space, which can be loosely interpreted as a form of hierarchical RL \cite{HierarchicalRL}. Each discrete bin corresponds to a region of the action space, within which a continuous action is learned. This method provides the benefits of discrete modes while allowing fine-grained control.

In the context of social robot navigation, multimodal behaviours often manifest in the angular velocity: a robot may turn left, turn right, or remain stationary while waiting for a human to pass. Based on this intuition, we discretize the angular velocity into three equal bins: left, straight, and right.

Formally, our model outputs a vector of dimension $3^H \times (2H+1)$. Each of the $3^H$ paths corresponds to a unique sequence of bin choices (e.g., middle $\rightarrow$ left $\rightarrow$ left). For each path, the model predicts a probability as well as a sequence of $2H$ continuous actions, which are squashed, scaled, and shifted to lie within the path's bins. The loss function combines the losses in Section \ref{sec:BC}:
\begin{equation}
L_\text{hybrid} = -w_d \log(p_i) + \sum_{j=1}^{3^H} p_j ||\mathbf{A}[j] - \mathbf{a}^*||_2^2 \text{.}
\end{equation}

Here, $i$ is index of the ground-truth discrete path, $w_d$ is a positive weight, and $\mathbf{a}^* \in \mathbb{R}^{2H}$ is the target action chunk. The second term is a weighted MSE for each predicted continuous action chunk $\mathbf{A}[j] \in \mathbb{R}^{2H}$. This loss is weighted by the model’s confidence $p_j$ in that path. This hybrid representation offers the multimodality of discrete spaces with the control precision of continuous actions. We find it provides a modest performance gain.

\subsection{Weight Perturbation}
An emerging challenge in continual learning is the ``loss of plasticity": the network's diminishing ability to learn as training progresses \cite{LossOfPlasticityInDeepContinualLearning}. In their study, \cite{LossOfPlasticityInDeepContinualLearning} trains a model on a sequence of binary classification tasks, where the target class pair changes over time. They observe that a network trained on prior tasks performs worse on the current task than one trained from scratch. This result is surprising because prior task training can be interpreted as pre-training. To preserve the model's plasticity, they use high weight decay and introduce random perturbations to the model weights.

A similar insight can be found in online RL: in \cite{BBF}, the policy network is periodically perturbed toward a randomly initialized network. While part of the performance gain may come from increased exploration when the model weights are perturbed, it is also plausible that weight perturbations help counteract the loss of plasticity as the policy's induced state-action distribution shifts over training.

Inspired by these findings, every 50k training steps, we instantiate a random network and move the current model's weights towards the random network by 30\%. Hence, $w_\text{model} \leftarrow 0.7 w_\text{model} + 0.3w_\text{random\_network}$. We gradually decay the size of these perturbations by $0.9$ every 50k training steps after 750k total training steps. Additionally, we first reset the weights to the best-performing checkpoint; this approach is inspired by evolutionary algorithms, where mutations are applied to the most fit individuals \cite{DE}.  We find that weight perturbation significantly enhances closed-loop performance.

\subsection{Reducing the Path-Tracking Latent Dimension}

The majority of our real-world dataset actions are simply path-tracking actions. This imbalance implies that a BC model that only attends to the path embedding could still achieve a relatively low loss without even considering $\StateHA$.

Models often prioritize feature `availability' over `predictivity' \cite{OnTheFoundationsOfShortcutLearning}. One notion of availability is the feature vector size: larger features can dominate the learned representation, even if they are not the most informative for the task \cite{OnTheFoundationsOfShortcutLearning}. Our empirical results show that when states are grouped by latent embeddings, they tend to cluster more strongly by path geometry than by robot–human configuration. While this does not necessarily prevent good action predictions, it suggests a potential over-reliance on path information.

To mitigate this potential issue, we reduce the dimensionality of the path-tracking embedding to 16 while leaving the human-avoidance embedding at 128. This limits the representational capacity available for encoding path information, encouraging the model to learn about human features. There is a modest improvement when employing this trick in both the real-world and simulation experiments.

\subsection{Traditional Machine Learning Parameters}
We find that our network and dataset require a small learning rate of $10^{-4}$ and approximately two million training steps. Attempts to reduce the number of gradient steps by increasing the learning rate resulted in poor performance.

As motivated by \cite{LossOfPlasticityInDeepContinualLearning}, we opt for a larger network with increased weight decay. Additionally, we find that using the AdamW optimizer is necessary for proper decoupled weight decay regularization \cite{AdamW}.

\section{Hardware Experiments}
\begin{figure}[t]
    \centering
    \includegraphics[width=0.5\textwidth]{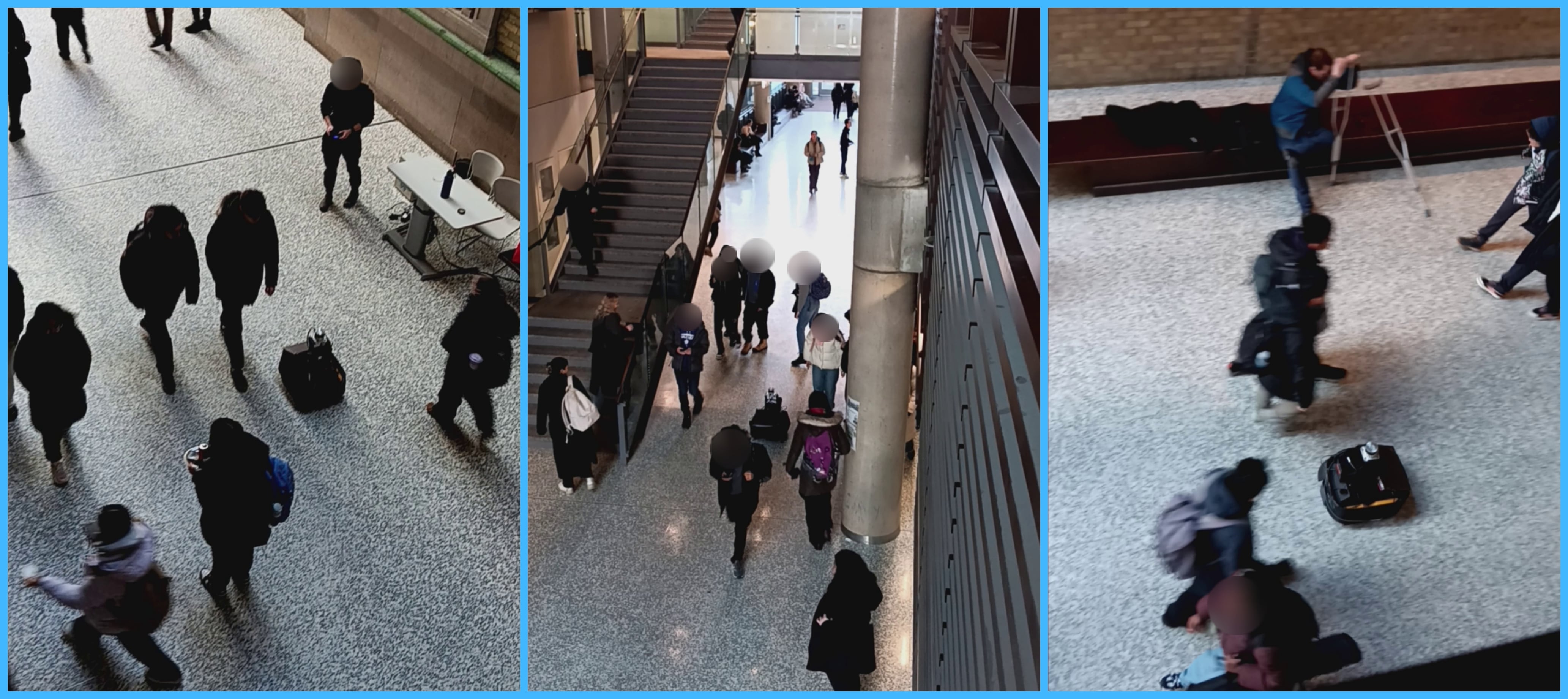}
    \caption{Examples of data collection on a university campus between class times, where the constrained spaces and high human density led to frequent robot–human interactions.}
    \label{fig:Data Collection}
\end{figure}
\subsection{Hardware Setup}
Our hardware platform closely follows \cite{DRMPC} (Figure \ref{fig:Full Pipeline}). We mounted an Ouster OS0-128 LiDAR on a Clearpath Jackal robot, which provides both range and reflectivity measurements. Leveraging Teach-and-Repeat (T\&R), the point cloud data supports robust localization, path retrieval, and rapid deployment in novel environments \cite{vtr}.

The LiDAR's high resolution allows us to perform human detection directly on the reflectivity image using a pre-trained YOLOX model (Figure~\ref{fig:RealWorldMoneyShot}) \cite{mmdetection}. Combining the range data with T\&R's localization, we track humans using Norfair \cite{norfair}.

For computation, T\&R runs on a Jetson Orin NX, while human tracking and model inference are executed on a ThinkPad P16 Gen~2. Communication between these devices and the Jackal platform is handled through ROS~2.

\begin{figure*}[t]
  \centering
  \includegraphics[width=0.95\textwidth]{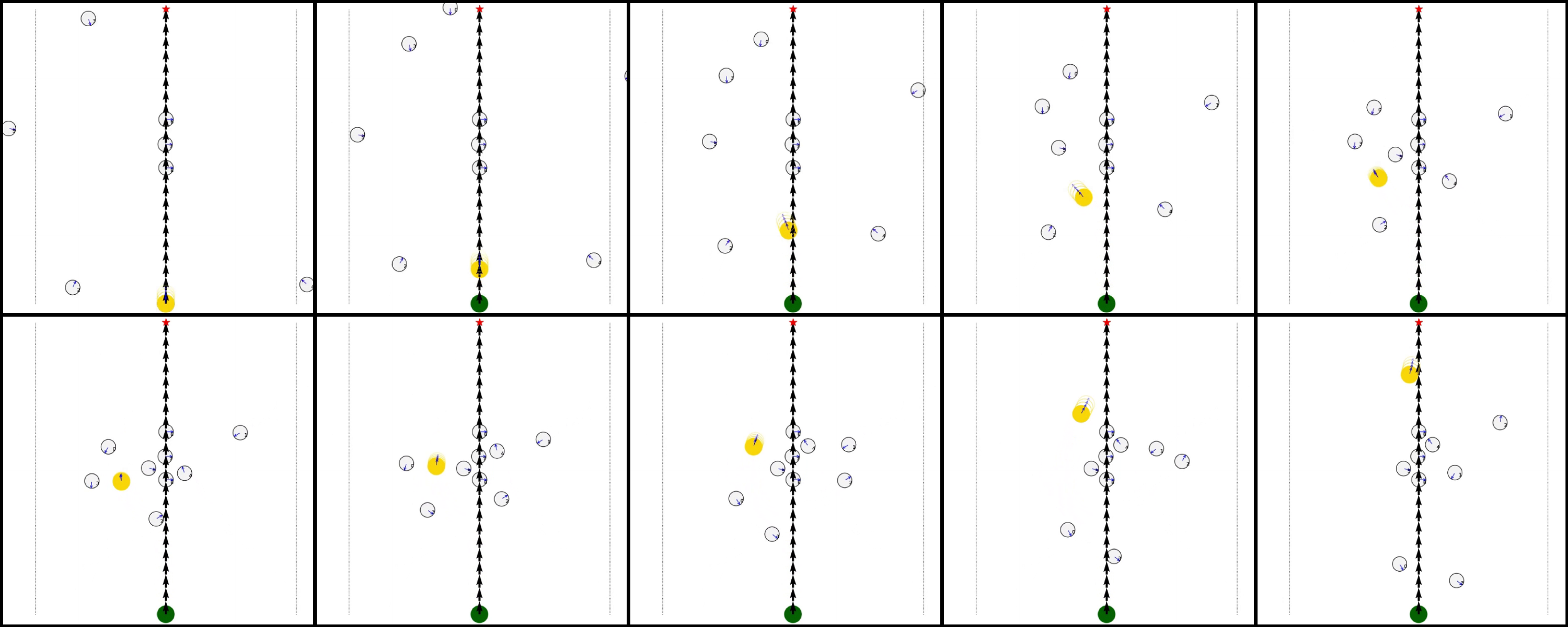}
  \caption{A visual example of a real-to-sim episode. The order of frames goes from left to right and top to down. We can see the robot (yellow) turning to avoid collisions (panels 3–4) with humans (grey) and slowing down (panels 5–7).
 }
  \label{fig:real to sim qualitative visual}
\end{figure*}

\begin{figure}[t]
    \includegraphics[width=0.45\textwidth]{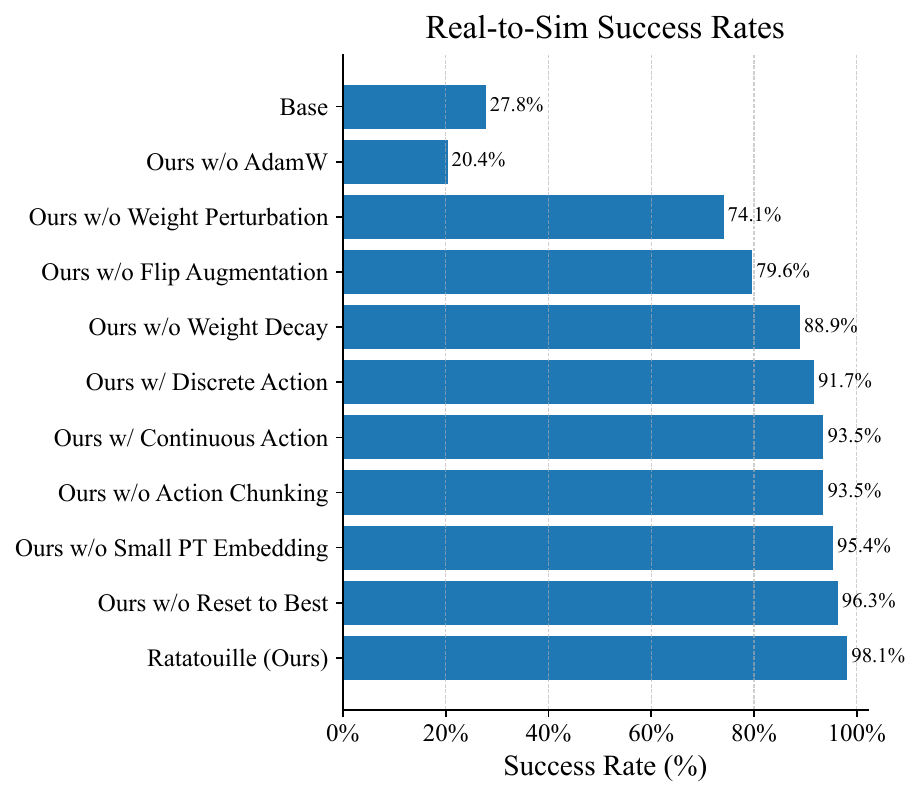}
    \caption{Real-to-sim success rates for Base, Ratatouille, and ablations of Ratatouille. The most substantial ingredients are AdamW for proper weight decay regularization, weight perturbation, and flip augmentation.}
    \label{fig:Real-to-sim success rates}
\end{figure}

 \begin{figure}[t]
  \centering
  \includegraphics[width=0.45\textwidth]{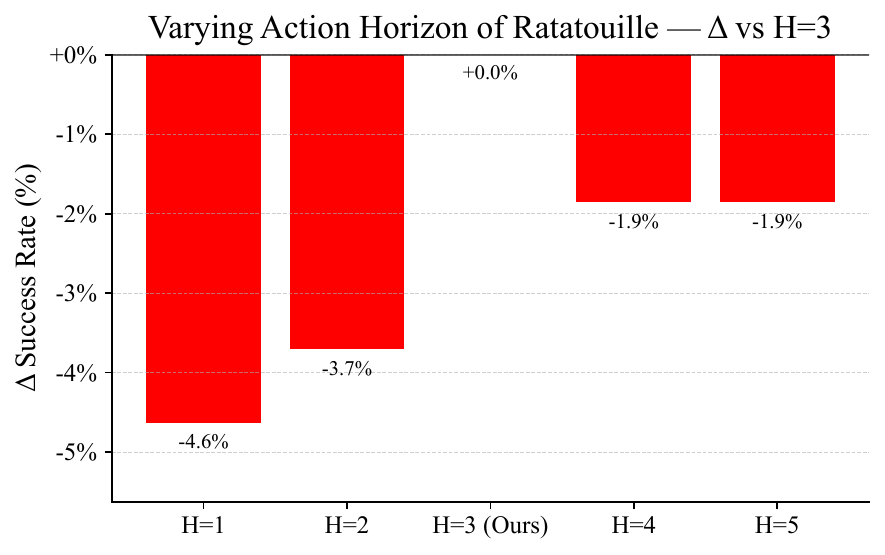}
  \caption{We vary $H$ for action chunking and observe the performance degradation relative to $H=3$ (Ours). Using a na\"ive $H=1$ results in the greatest performance degradation.}
  \label{fig:action_chunking_ablation}
\end{figure}
\subsection{Data Collection}


By default, the robot autonomously follows the reference path using T\&R’s Model Predictive Control (MPC) controller, while the expert human operator can override the MPC actions when necessary. We collected 11 hours of data (400k environment steps) over two weeks on a university campus. To maximize the diversity of human-robot interactions, we prioritized data collection during peak pedestrian traffic—such as the transition periods between classes (Figure \ref{fig:Data Collection}). These times typically range between 5 to 30 people around the robot. We also collected lower-density scenes of 0 to 4 people.

This in-the-wild data collection resulted in significantly more diverse interactions than those typically found in simulation \cite{DRMPC}. For example, we encountered pedestrians intentionally obstructing the robot’s path, groups walking alongside the robot, clusters of people engaged in conversation, and individuals distracted while using their phones. Despite the rich spectrum of robot-human interactions, our real-world dataset is imbalanced, with 73\% of the demonstrations reflecting the robot simply following T\&R’s MPC controller. 

\begin{figure}[t]
  \centering
  \includegraphics[width=0.4\textwidth]{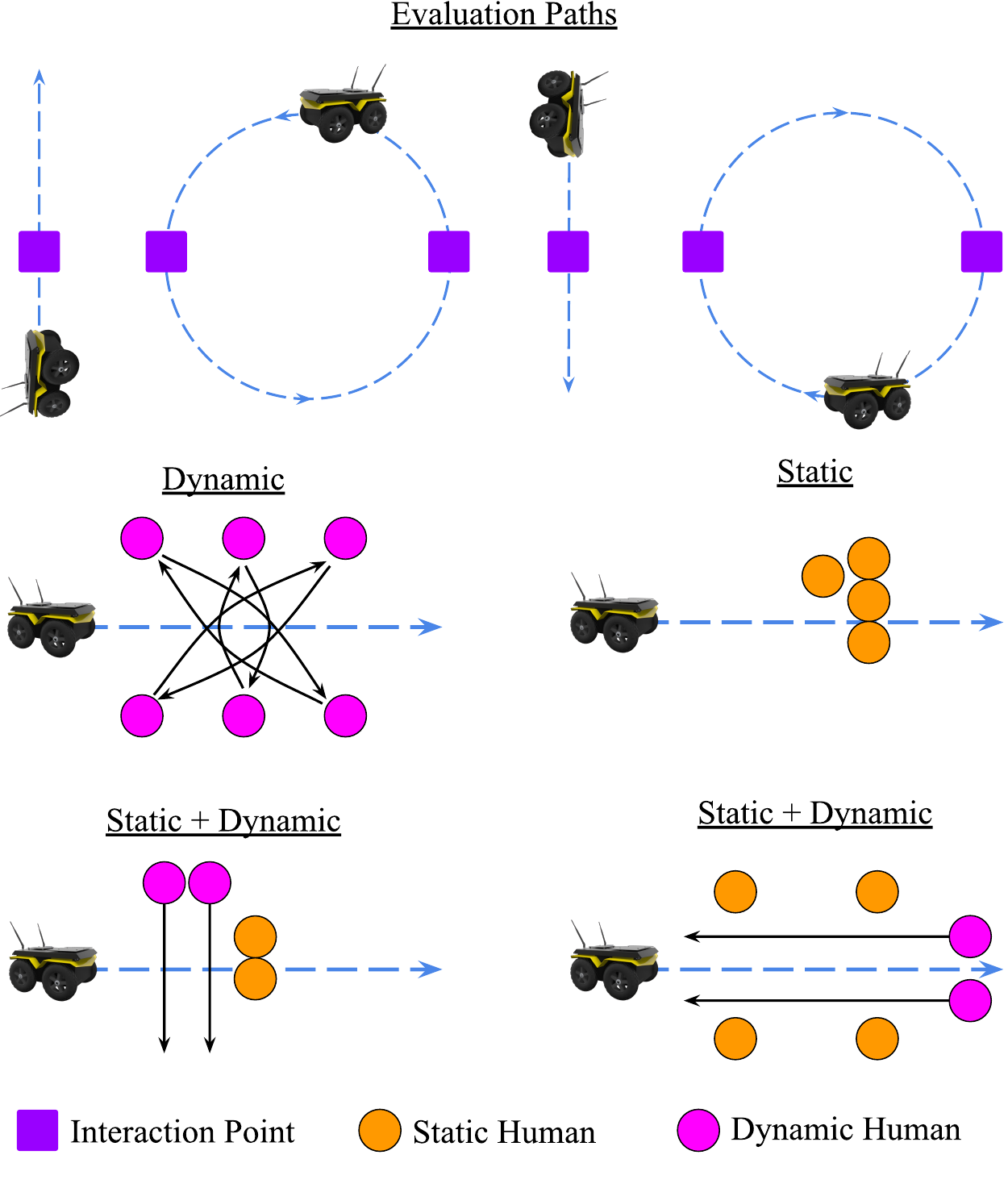}
  \caption{The paths and examples of human configuration scenarios for real-world quantitative evaluation. The example interactions in this figure are not exhaustive.}
  \label{fig:real-world scenarios}
\end{figure}

\subsection{Real-to-Sim Experimental Setup}\label{sec:real_to_sim_experimental_setup}

We first quantitatively evaluate our model trained on real-world data in simulation. We then present quantitative results from controlled real-world experiments, followed by a qualitative evaluation in a public food court during lunch hour.

We use the simulator from \cite{DRMPC} that combined path tracking and human avoidance. In this simulator, and many others, humans are modeled using Optimal Reciprocal Collision Avoidance (ORCA) \cite{ORCA}. Our evaluation suite consists of 27 unique scenarios, constructed by combining different paths with different human configurations. The three paths include a straight 8-meter path, a CW circular path with an 8-meter diameter, and a CCW circular path of the same size.

For each path, we test the robot's behaviour under different levels of human presence. Specifically, we vary the number of static humans directly on the path between one, two, and three. Separately, we introduce either zero, three, or six dynamic humans moving through the scene. For these dynamic pedestrians, we maintain a 2:1 ratio of `regular' to `aggressive' behaviour profiles. Aggressive humans are intended to simulate distracted individuals—such as those using their phones—or those who intentionally challenge the robot. To model this, we greatly reduce ORCA's safety margin. 

Although there is a real-to-sim gap, simulation enables large-scale quantitative evaluation. In this gap, simulated humans follow different movement patterns compared to humans in the real world. Moreover, the simulation lacks realistic sensor noise and processing delays. While these differences might seem advantageous, from a ML perspective, they may be harmful. These differences introduce a distributional mismatch between the training and evaluation conditions, which may affect performance in unpredictable ways. 

Despite these mismatches, simulation allows for evaluation under consistent and repeatable conditions across diverse scenarios. Conducting such evaluations in the real world would require significant effort and coordination, particularly when comparing across multiple model ablations. Additionally, it would be nearly impossible to reproduce the same human behaviour for each trial—humans would inevitably respond differently when interacting with the first versus the last model tested. For these reasons, simulation provides a practical and scalable option for holistic evaluation, which complements the real-to-real and sim-to-sim experiments that follow.

We evaluate a total of 11 models. We begin with the `Base' model, which is na\"ive BC without any ingredients. We then evaluate Ratatouille, which has all the ingredients. Then, we ablate every ingredient. To determine the effect of the hybrid action space, we test a `Continuous' model that predicts the continuous action chunk, and we test a `Discrete' model that uses a discrete action space of 25 bins with an action horizon of one to maintain statistical efficiency. Both the Continuous and Discrete model have all other ingredients active. 
    
We also ablate the heuristic of resetting to the best model before applying the perturbations. The reduced path-tracking embedding size is 16, and for the ablation we increase it to 128 to match the human-avoidance embedding size. For the optimizer ablation, we replace AdamW with Adam. The weight decay ablation sets the weight decay to zero instead of the default value of $5 \times 10^{-3}$. Lastly, in the action chunking ablation, the action horizon is reduced from three to one.

\subsection{Real-to-Sim Evaluation}

The average success rate across 108 episodes is presented in Figure \ref{fig:Real-to-sim success rates}. The performance gap between the Base model and Ratatouille is substantial: the Base model achieves a success rate of 27.8\%, while Ratatouille reaches 98.1\%. 

The impact of each ingredient varies in magnitude. Notably, without AdamW, weight decay regularization is not decoupled from the gradient calculation of the loss function. We found its large effect surprising, suggesting that further investigation into AdamW’s role in other IL problems is warranted.

We next observe that weight perturbation yields a substantial performance gain, despite prior work demonstrating its benefits primarily in continual learning. Although BC is typically cast as a standard supervised learning problem, we hypothesize that its advantage here stems from the inherently multi-objective nature of social robot navigation, where the policy simultaneously learns human avoidance and path tracking, with the relative emphasis on each task depending on the state. This stands in contrast to traditional classification tasks, where the learning objective remains consistent across the dataset.

By injecting variability, weight perturbation may help preserve the network’s capacity to adapt to both objectives. This is consistent with \cite{LossOfPlasticityInDeepContinualLearning}, which demonstrates that networks lose plasticity when repeatedly trained on changing targets. If BC in social navigation is interpreted as multi-task learning—where task emphasis fluctuates across minibatches—then weight perturbation serves to mitigate the loss of plasticity.

Validating this hypothesis would require either (i) testing weight reset ablations on a comparable multi-objective task, or (ii) training with a batch size equal to the entire dataset to eliminate variation in task distribution across minibatches. The latter is computationally demanding, as it requires gradient computation over the full dataset at each step and careful memory management to accommodate GPU limits. We leave such analyses for future work. We do provide more extensive evidence in Section~\ref{sec:Simulation Experiments}, showing that weight perturbation consistently improves performance in social robot navigation.

Given the importance of weight perturbation, it is not surprising that weight decay also has a strong effect on performance. In \cite{LossOfPlasticityInDeepContinualLearning}, both weight perturbations and weight decay were found to be necessary to preserve a network’s ability to learn, further supporting our hypothesis. 

We also observe that the hybrid action head contributes positively, but, unexpectedly, the continuous action representation outperforms the discrete one. This result is surprising because with the flip augmentation applied, the dataset is guaranteed to be multimodal. Despite this, the continuous action representation achieves a strong performance of 93.5\%.

Next, we find that both using a smaller path-tracking embedding and resetting to the best-performing weights during training yield only minor improvements. For action chunking, we perform a standard ablation by varying the horizon $H$ (Figure~\ref{fig:action_chunking_ablation}). The results exhibit an inverted-U shape: performance peaks at $H=3$, though the overall impact of action chunking remains modest compared to other ingredients.

Finally, we present the qualitative behaviour of Ratatouille in simulation. Figure~\ref{fig:real to sim qualitative visual} shows an episode involving nine humans, where the robot’s next three actions are visualized. The model slows down appropriately in cluttered regions and turns to avoid collisions, while still making progress toward its goal.

\begin{table}[htbp]
\centering

\setlength{\tabcolsep}{6pt}
\caption{\small Quantitative real-world results averaged across 24 episodes. The expert serves to provide an intuitive benchmark, but it is not used in the ranking of the learned models.}
\begin{tabular}{lccc} 
\toprule
\textbf{Model} & \textbf{SR ($\uparrow$)} & \textbf{CPM ($\downarrow$)} & \textbf{NT ($\downarrow$)} \\
\midrule
Base                & 0.38 & 0.06 & \textbf{23.3} \\
Ratatouille (Ours)  & \textbf{0.79} & \textbf{0.01} & 28.2 \\
SCAND \cite{SCAND}              & 0.25 & 0.08 & 37.5 \\
Expert        & 1.00 & 0.00 & 25.2 \\
\bottomrule
\end{tabular}

\label{tab:real world}
\end{table}

\subsection{Real-to-Real Experimental Setup}
We next evaluate our model quantitatively in the real world. Figure~\ref{fig:real-world scenarios} presents the evaluation paths and a few examples of the human configurations tested. We ensured diversity in these configurations, including purely dynamic, purely static, and mixed dynamic–static scenarios. Each scenario involved between four and six humans interacting with the robot.

We evaluate four models. The first two are the \emph{Base} model and \emph{Ratatouille}. The third benchmark is \emph{SCAND} \cite{SCAND}, which performs BC using a BEV image of the LiDAR data as $\StateHA$. We include SCAND both because it is a recent BC approach for social robot navigation and because it provides a useful contrast in state representation. Lastly, we evaluate the \emph{Expert} model, where a human teleoperator controls the robot, who is the same expert who collected the training data.

In total, we evaluate 24 episodes: six runs on each evaluation path, yielding 36 human–robot scenarios. The evaluation paths closely match those in simulation, with a radius of 4.5\,m. We report results using success rate (SR), collisions per meter (CPM), and navigation time (NT) per episode. We deliberately avoid ambiguous metrics such as distance to the path or to humans. It is often necessary for the robot to deviate from its path to avoid humans. Proximity to a human does not imply poor performance; for instance, a human walking alongside the robot represents acceptable behaviour.

\vspace{-3mm}
\subsection{Real-to-Real Evaluation}
The results are presented in Table~\ref{tab:real world}. Ratatouille achieves a substantially higher SR and a sixfold reduction in CPM compared to the Base model. The Base model has the lowest NT, but this is largely because it often drives directly into humans. In such cases, we pause the evaluation, allow the human in the collision to move out of the way, and then resume. Without this intervention, the Base model would often fail to complete the episode. Overall, the relative results between the Base model and Ratatouille align well with the real-to-sim findings.

It is also worth noting that although Ratatouille has a non-zero CPM, its collisions are qualitatively more acceptable: they often occur when Ratatouille is turning and clips a human slightly, which is preferable to the full head-on collisions that the Base model and SCAND frequently encounter.

SCAND, on the other hand, is generally able to follow the path and reach the end of an episode, but it moves slower than all other models. It also achieves the lowest SR and the highest CPM. The relatively poor performance of SCAND compared to the Base model likely stems from its less efficient state representation: SCAND relies on a bird's-eye view of the LiDAR point cloud, whereas our compact state representation captures only the most essential components of the scene.

\vspace{-3mm}
\subsection{Food Court Qualitative Evaluation}

We also evaluated Ratatouille in a public food court during peak lunch hours. Qualitatively, we observed several proficient behaviours for social robot navigation. For instance, the robot was able to smoothly turn around static humans, navigate fluidly among crowds walking in the same direction, and slow down in cluttered situations. Nonetheless, occasional collisions still occurred—for example, when a human abruptly jumped in front of the robot or when the robot clipped a human while attempting to turn around them. A video of this evaluation can be found \url{https://youtu.be/tOdLTXsaYLQ}.

\vspace{-3mm}
\section{Simulation Experiments}\label{sec:Simulation Experiments}

\begin{figure}[t]
  \centering
  \includegraphics[width=0.45\textwidth]{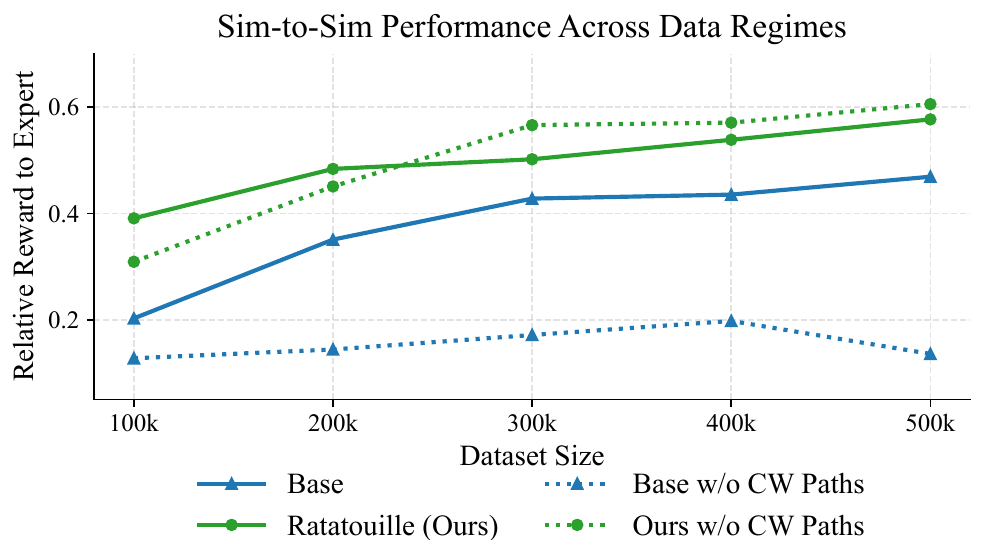}
  \caption{Ratatouille outperforms the Base model across all data regimes. Ratatouille's performance does not degrade if CW paths are removed from training due to flip augmentation.}
  \label{fig:FA analysis}
\end{figure}

\begin{figure}[t]
  \centering
  \includegraphics[width=0.45\textwidth]{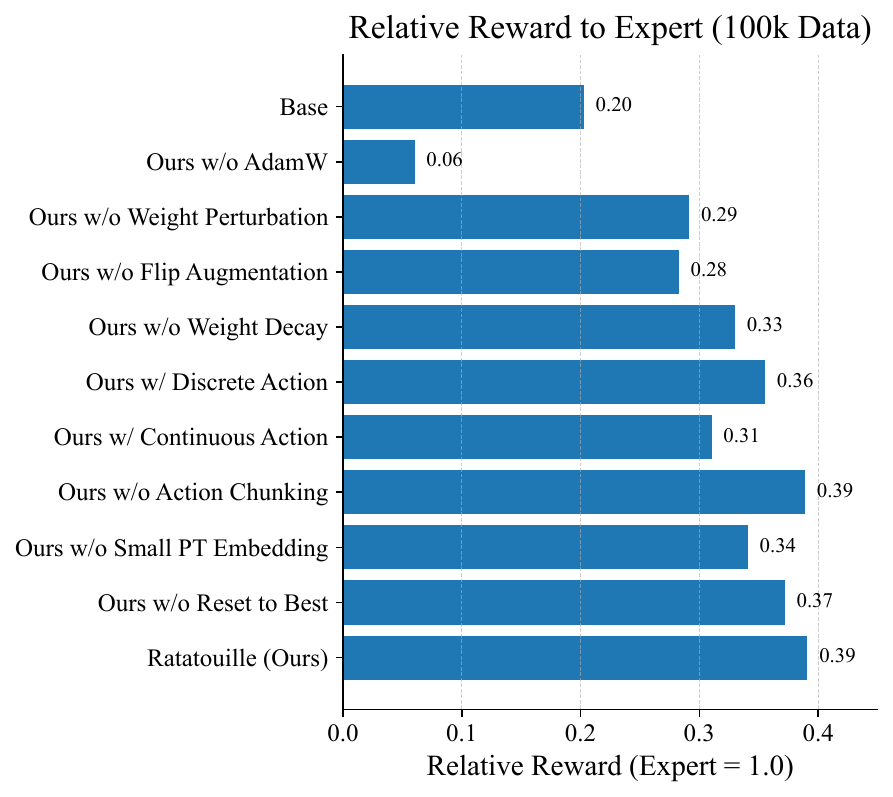}
  \caption{Ablation of Ratatouille in the simulation 100k setting. The order of plotting is the same as Figure \ref{fig:Real-to-sim success rates}. All ingredients except action chunking provide a benefit.}
  \label{fig:Ratatouille Ablation in Simulation}
\end{figure}

\begin{figure}[t]
  \centering
  \includegraphics[width=0.4\textwidth]{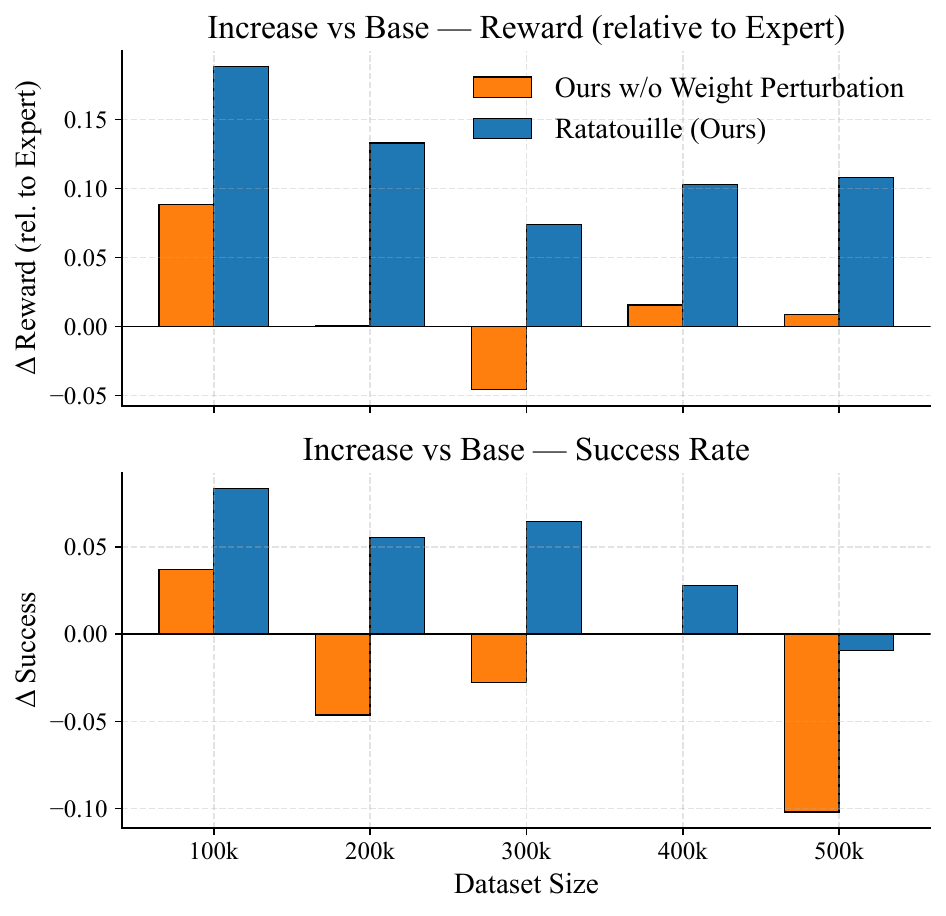}
  \caption{Weight perturbation in simulation experiments always aids performance. The $y$-axis is $\Delta$ to the Base model.
}
  \label{fig:Weight Perturbation Sim}
\end{figure}

We replicate the hardware experiments in simulation where the reward function is known. An expert policy is trained with DRL, and its collision-free rollouts form the demonstrations for IL. If Ratatouille achieves the highest reward in this setup, it provides strong evidence that training on real-world data leads to maximizing the expert’s underlying reward function, which implicitly encodes the semantics of human social navigation. Accordingly, the primary metric in this section is the normalized reward relative to the expert. The simulator and evaluation setup are identical to those in Section~\ref{sec:real_to_sim_experimental_setup}. The reward function follows \cite{DRMPC}.

\vspace{-5mm}
\subsection{Sim-to-Sim Evaluation}

From Figure~\ref{fig:FA analysis}, we observe that increasing data for Ratatouille consistently improves performance, and Ratatouille always outperforms the Base model. To analyze the effect of flip augmentation, we remove CW paths from the training data while keeping the evaluation scenarios fixed. Because flip augmentation increases path diversity, Ratatouille with and without CW paths performs nearly identically. In contrast, the Base model exhibits a sharp performance drop when trained without CW paths, indicating poor generalization to CW trajectories at test time.

From Figure~\ref{fig:Ratatouille Ablation in Simulation}, we observe results that closely mirror the real-to-sim evaluation: Ratatouille achieves the highest reward, though the relative importance of each component shifts slightly. AdamW, weight perturbation, and flip augmentation remain the three most impactful ingredients. Action chunking, however, shows little effect, likely because the simulated expert, generated by a DRL policy, exhibits largely Markovian behaviour. Action chunking is expected to be more beneficial when expert demonstrations exhibit non-Markovian behaviour, as is often the case in real-world data collection \cite{QChunking}.

Examining Figure~\ref{fig:Weight Perturbation Sim}, we find that across all data regimes, weight perturbation always improves both reward and success rate. We note that at 500k data samples, Ratatouille achieves a higher reward but a lower success rate compared to the Base model. While a high success rate is a necessary condition for high reward, it is not sufficient: in these simulation experiments, the primary metric of interest is the reward, whereas success rate serves as a more intuitive but secondary indicator.

\section{Limitations and Future Work}

While our hardware setup is reliable, it remains limited compared to the expert's observational capabilities. The expert can infer human intent from subtle body cues and benefits from perfect perception without delays, whereas the robot only observes human positions. A potential solution is to use virtual reality goggles to restrict the demonstrator to the robot's view.

One ongoing line of work is investigating the effect of weight perturbations in other IL problems. If our hypothesis holds, this simple technique could yield large performance gains, representing a substantial contribution to IL.

Our flip augmentation scheme played a critical role in Ratatouille's performance. A natural extension would be to design a model architecture that is inherently invariant to reflections (`reflectivariance'). Beyond flips, another exciting avenue is to leverage generative models to produce realistic augmented data, further improving robustness and generalization.

\vspace{-3mm}
\section{Conclusion}
We investigated the limitations of na\"ive BC for social robot navigation and showed that strong performance depends on a careful combination of architectural and training ingredients. Our proposed pipeline, Ratatouille, improves safety and reliability without additional data, reducing collisions per meter by $6\times$ and tripling success rate. Through extensive ablations in both simulation and real-world deployments, we demonstrated the necessity of each ingredient, underscoring that principled IL design is critical for safe and reliable social robot navigation.

\vspace{-3mm}
\section*{Acknowledgment}
We thank the MaRS Discovery District for allowing us to conduct qualitative evaluations in their food court.

\bibliographystyle{IEEEtran}
\bibliography{ref}

\appendices

\end{document}